\documentclass[fleqn,10pt]{wlpeerj}

\usepackage{multirow}
\usepackage{hyperref}

\title{MEP: Multiple Kernel Learning Enhancing Relative Positional Encoding Length Extrapolation}

\author[1]{Weiguo Gao}
\affil[1]{independent researcher}
\corrauthor[1]{Weiguo Gao}{gaoweiguo801@gmail.com}

\begin{abstract}
When the predicted sequence length exceeds the length seen during training, the transformer's inference accuracy diminishes. To address these challenges, the ability to generalize or extrapolate to longer sequence lengths has garnered increased attention. Existing relative position encoding methods, such as those based on the ALiBi technique, address the length extrapolation challenge exclusively through the implementation of a single kernel function, which introduces a constant bias to every post-softmax attention scores according to their distance. These approaches do not investigate or employ multiple kernel functions to address the extrapolation challenge. Drawing on the ALiBi approach, this study proposes a novel relative positional encoding method, called MEP, which employs a weighted average to combine distinct kernel functions(such as the exponential kernel and the Gaussian kernel) to generate a bias that is applied to post-softmax attention scores. Initially, the framework utilizes various kernel functions to construct multiple kernel functions. Each kernel function adheres to a consistent mean weight coefficient, harnessing the synergistic advantages of different kernels to formulate an innovative bias function. Subsequently, specific slopes are tailored for each kernel function, applying penalties at varying rates, to enhance the model's extrapolation capabilities. Finally, this bias is seamlessly incorporated as a penalty to the post-softmax scores. We present two distinct versions of our method: a parameter-free variant that requires no new learnable parameters, which enhances length extrapolation capabilities without compromising training efficiency, and a parameterized variant capable of integrating state-of-the-art techniques. Empirical evaluations across diverse datasets have demonstrated that both variants of our method achieve state-of-the-art performance, outperforming traditional parameter-free and parameterized approaches.
\end{abstract}

\begin{document}

\flushbottom
\maketitle
\thispagestyle{empty}
\section*{Introduction}
The Transformer-based language model~\citep{vaswani2017attention,dufter_position_2022} has seen widespread application in a variety of natural language processing tasks, achieving state-of-the-art (SOTA) performance in domains including  language modeling~\citep{devlin_bert_2019}, code completion~\citep{chen2021evaluating}, and text summarization~\citep{zhang2020pegasus}, among others. However, due to the Transformer architecture's computational complexity, which scales quadratically with input length ($\mathcal{O}(L^2)$), training is generally limited to shorter text sequences\citep{chi2022kerple}, with token limits frequently set at 1024~\citep{zhang2020pegasus}, or 512~\citep{raffel2020exploring}. This constraint poses a significant challenge for predictions extending beyond the maximum training lengths, often leading to a reduction in the model's inferential accuracy. To address these issues, the ability of models to generalize or extrapolate to longer sequence lengths has gained increased attention. Specifically, this refers to the model's capacity to accurately predict sequences that extend beyond the lengths encountered during training, an ability known as length extrapolation~\citep{press2021trainalibi}.

The Transformer model utilizes positional encoding to capture the sequential order of input sentences. Several forms of positional encoding exist. Absolute Positional Encodings(APE)~\citep{vaswani2017attention}, which utilize sine and cosine functions, are inadequate for length extrapolation. Relative Positional Encoding(RPE) ~\citep{su2024roformer,raffel2020exploring,press2021trainalibi,chi2022kerple,li_functional_2023_fire,peng_yarn_2023,jacot_neural_2018_ntk} represents an alternative approach, encompassing methods such as Rotary Positional Encoding (RoPE) ~\citep{su2024roformer} and ALiBi ~\citep{press2021trainalibi}. These methods have been broadly employed in large language models (LLMs)~\citep{openai_gpt-4_2023,touvron_llama_2023}, forming the basis for the optimization and enhancement of various positional encoding strategies to augment length extrapolation capabilities. 

\begin{table*}
\centering
\begin{tabular}{ll|ccccc}
\toprule
 Type&PE& Manifestation& Learnable & Kernel Function &Formula\\
\midrule
 {APE}  &Sinusoidal & Embedding & Yes &-&-\\
\midrule
 {RoPE-type}  &{RoPE}    & Embedding & No &- &-\\
 &{YaRN}     & Embedding & No     &- &-\\
\midrule
 {ALiBi-type} &T5 Bias   & Bias      & Yes &Learning &-\\
 &{ALiBi}    & Bias      & No &Exponential & ${\exp(-|j-i|)}$\\
 &{SANDWICH} & Bias      & No &Exponential & $\exp(p_i^Tp_j)$\\
 &{KERPLE}   & Bias      & Yes &Polynomial & ${(1 + r_2 |j-i|)^{-r_1}}$\\
 &     &   &  &Gaussian-like & $\exp({r_1} \|{j} -{i}\|^{r_2})$\\
 &{FIRE}     & Bias      & Yes &Learning &-\\
 &{MEP}     & Bias      & No/Yes &MKL &MKL\\
\bottomrule
\end{tabular}
\caption{A list of extrapolatable PEs. \emph{Manifestation}: whether a PE encodes position information in embeddings or by directly modifying the post-softmax attention mechanism. \emph{Learnable}: whether a PE is learnable or not, which determines whether it can adapt and adjust based on the input. \emph{Kernel Function}: Which function is employed as bias in alibi-type positional encoding (PE)? \emph{ formula}: specific formula in kernel functions. MKL means Multiple Kernel Learning. The variables i/j denote the i-th or j-th position in a sequence. $|j-i|$ represents the relative distance between positions i and j, while $p_i$ and $p_j$ denote position embedding at those positions, respectively. Exp denotes the Exponential kernel function, and $r_1$ and $r_2$ are the corresponding parameters.}
\label{tab:pes}
\end{table*}

\begin{figure}
    \centering
    \includegraphics[width=\linewidth]{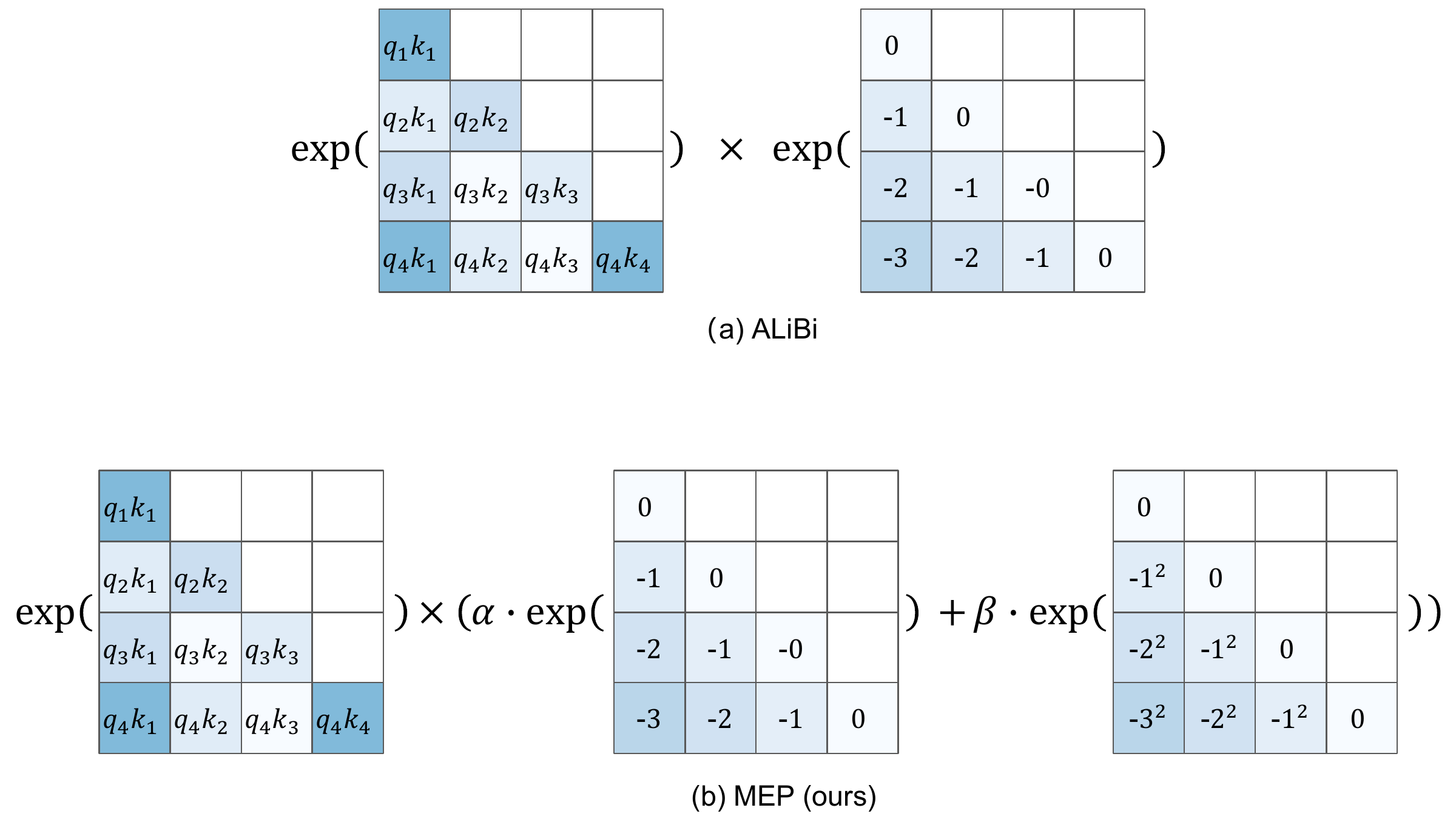}
    \caption{(a) Previous positional encoding ALiBi produces a single exponential kernel function to post-softmax attention scores. For a transformer language model with $H$ attention heads, the range of $h$ is $n\cdot\frac{8}{H}$, where $n=\{1\dots H\}$. Left = the post-softmax self-attention matrix, right = the temporal biases matrix. (b) In contrast, the proposed MEP positional encoding builds a bias by multiple kernel functions to every post-softmax attention scores according to their distance. We employ multiple kernel learning, merging exponential, Gaussian, and polynomial kernels. In the exponential and Gaussian kernels, the range of $h$ aligns with that of ALiBi, while in the polynomial kernel, $h$ represents learned parameters. Left = the post-softmax self-attention matrix, middle = the exponential kernel temporal biases matrix, right = the Gaussian kernel temporal biases matrix. for example, $\alpha=0.5$ and $\beta=0.5$ is coeffient. exp denotes the Exponential kernel. $\text{slopes value=1}$.} 
    \label{fig:alibi-mep}
\end{figure}

\begin{figure}
    \centering
    \includegraphics[width=\linewidth]{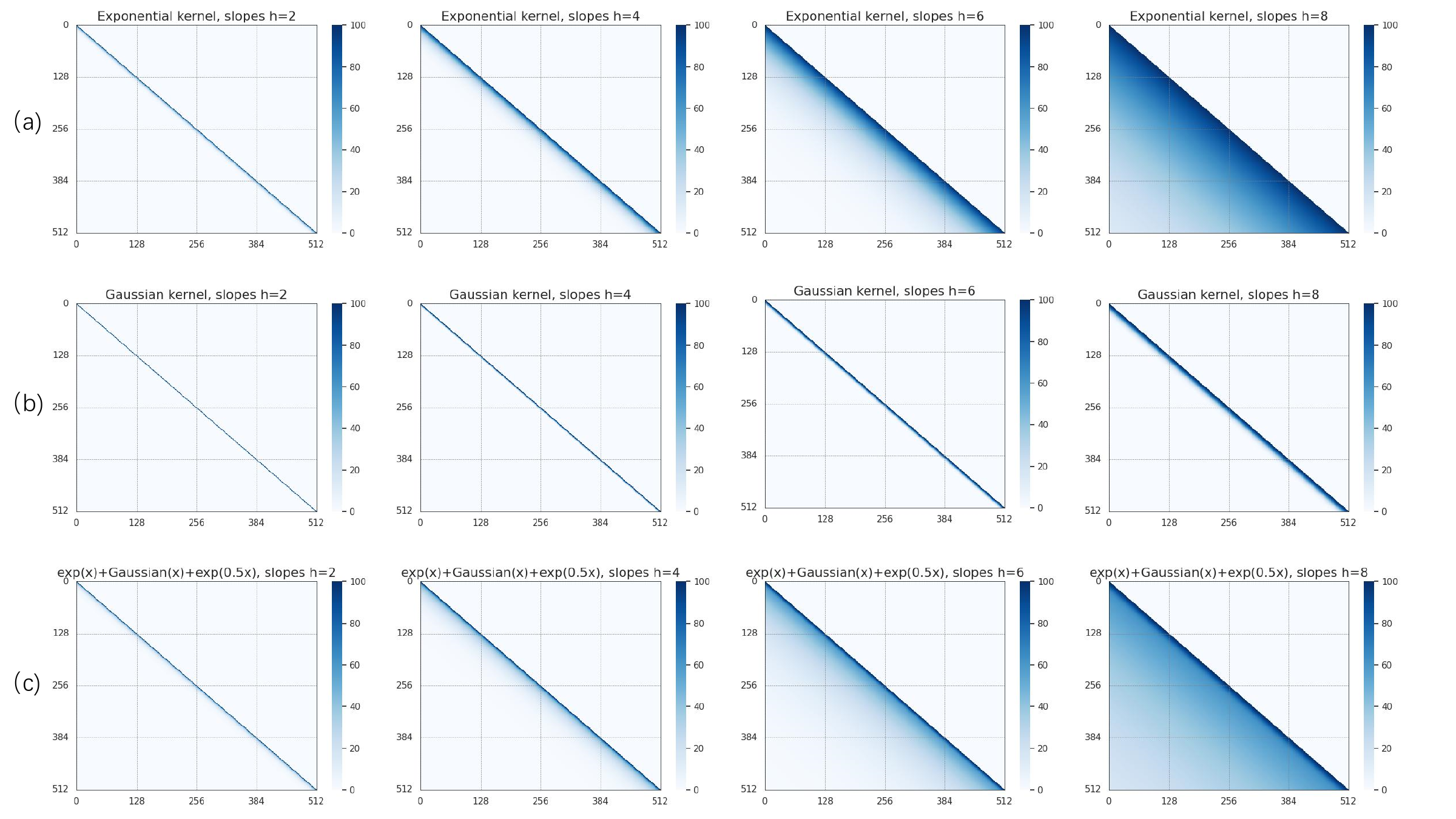}
    \caption{Each point denotes the post-softmax attention score corresponding to the relative position $|i-j|$, obtained after passing through the kernel function. (a) Exponential kernel post-softmax attention scores, head = 2 to 8. (b) Gaussian kernel scores. (c) ours(MEP parameter-free model) MKL’s scores.} 
    \label{fig:alibi-gaussian-postsoftmax-attentions}
\end{figure}

There exist two primary categories of Relative Positional Encoding(Table \ref{tab:pes}): the \textbf{RoPE-type relative position encoding} and the \textbf{ALiBi-type relative position encoding}. \textbf{The RoPE-type  RPE}: These methods represent enhancements over the baseline RoPE model, employing high-frequency extrapolation and low-frequency interpolation~\citep{peng_yarn_2023}. Recent studies\citep{kazemnejad2024impact,press2021trainalibi} have revealed that RoPE does not generalize effectively to contexts longer than those encountered during training. In contrast, the YaRN~\citep{peng_yarn_2023} method—a RoPE modification that integrates attention scaling and NTK-by-parts~\citep{jacot_neural_2018_ntk,peng_yarn_2023} interpolation—has demonstrated superior performance, surpassing all previous NTK-aware interpolation methods in scenarios with and without fine-tuning. \textbf{ALiBi-type RPE}: Similarly, the methods constitute improvements upon the ALiBi~\citep{press2021trainalibi} framework, involving the modification of the bias function to enhance extrapolation capabilities. Figure~\ref{fig:alibi-mep}(a), ALiBi positional encoding employs a single kernel function, utilizing an exponential kernel ($\exp(-|i-j|)$) ~\citep{gonen2011multiple}, and incorporates bias based on the pairwise distances into the post-softmax attention scores(softmax normalization process within the self-attention mechanism~\citep{vaswani2017attention}), details see Eq.~\eqref{eq:postsoftmaxA}. However, the function rapidly approaches the zero point~\citep{chi2022kerple}.  While characterized by its straightforward implementation, it has been surpassed in performance terms. KERPLE~\citep{chi2022kerple}, a framework aimed at improving relative position embedding for extrapolation, leverages kernelized positional differences and introduces two distinct bias kernel functions—polynomial and Gaussian-like. This approach does not investigate the synergistic effects of integrating two kernel functions. However, this incorporation of additional trainable parameters results in diminished training velocities. Sandwich~\citep{chi_dissecting_2023} is a pioneering parameter-free approach in relative positional embedding that relies exclusively on the inner product of two position embeddings. While it forgoes the addition of trainable parameters, its performance is somewhat limited. It bears resemblance to an exponential kernel function($\exp(p_m*p_n)$), with the key difference being that its exponential term is denoted by ($p_m*p_n$). T5's approach~\citep{raffel2020exploring} to relative positional encoding achieves generalization to longer contexts by employing a consistent positional representation for out-of-distribution (OOD) sequence lengths. However, this method encounters latency in vector operations on contemporary accelerators when processing longer sequences. FIRE~\citep{li_functional_2023_fire} employs a learnable continuous function and addresses inputs beyond the training domain with progressive interpolation, encoding positions of any sequence length within a 0/1 range. However, the addition of parameters slows the process.

This article focuses on enhancing the relative position encoding within the ALiBi-type RPE. The objective is to impose a penalty on the post-softmax attention scores following softmax normalization. The nearer the distance, the higher attention score, and conversely, the greater the distance, the lower attention score. ALiBi, KERPLE, and Sandwich each address the challenge of length extrapolation through a singular kernel function, incorporates bias based on the pairwise distances into the post-softmax normalization process within the self-attention mechanism, whilst overlooking the potential advantages of harnessing multiple kernel functions for extrapolation. A single kernel function, such as the exponential, polynomial, or Gaussian kernel may not be sufficient to capture all the nuances of the data~\citep{gonen2011multiple,sonnenburg2006largeMKL}. Can one synergize various positional encodings? What is the methodology for their integration?  Multiple kernel learning (MKL)~\citep{gonen2011multiple,sonnenburg2006largeMKL} aims to overcome this limitation by employing a combination of kernel functions instead of relying on a single one. The concept involves leveraging the complementary strengths of different kernels, which are typically amalgamated in a weighted sum. As illustrated in Figure~\ref{fig:alibi-mep}(b), 1) Guided by the concept of MKL, expanding on the ALiBi framework, this paper introduces a novel relative positional encoding strategy called \textbf{MEP} (\textbf{M}ultiple Kernel Learning \textbf{E}nhancing Relative \textbf{P}ositional Encoding Length Extrapolation), which integrates multiple kernel functions, such as the exponential, Gaussian, and polynomial kernels, with each kernel assigned an average weight coefficient. We propose two fusing methods: a parameter-free variant that requires no new learnable parameters and utilizes both exponential and Gaussian kernel functions for merging, as well as a parameterized variant capable of combining exponential and KERPLE polynomial kernel functions with learnable parameters. 2) Slopes play a crucial role in length extrapolation. For various kernel functions, distinct slopes are utilized to enhance the model's extrapolation performance. For example, the exponential and Gaussian kernels use slopes from ALiBi(For models with 8 heads, the slope values are: $\frac{1}{2^1}, \frac{1}{2^2}, \dots, \frac{1}{2^8}$.), while the polynomial kernel employs learned slopes. 3)The final bias is calculated through cumulative summation, where each bias is the product of a weighted average coefficient and the each kernel function result (the kernel inputs include the slope value and the i-j distance). These bias is seamlessly integrated as a penalty into post-softmax attention scores, effectively modulating penalties at diverse rates. Our approach MEP enhances length extrapolation performance, and maintains training efficiency. Empirical testing across a variety of datasets demonstrates that our method secures state-of-the-art (SOTA) outcomes in comparison with parameter-free and parameterized approaches.

Contributions of this paper are as follows:
\begin{enumerate}
    \item This paper proposes a novel ALiBi-type RPE, which fusing multiple kernel to get bias as penalty to post-softmax socres. Our method is simple and can be seamlessly integrated with any alibi-type positional encoding method.
    \item Our approach not only achieves state-of-the-art (SOTA) performance in comparison to non-learnable and learnable relative positional encoding methods but also demonstrates robust performance across various datasets.
    \item We provide a theoretical demonstration of our method's superiority over both ALiBi.
    \item We delve into the analysis of our method's effectiveness through comprehensive ablation experiments. Detailed experimental investigations are conducted on the selection of slope parameters, analyzing their influence on length extrapolation. We underscore the criticality of slope selection and offer practical recommendations for kernel function optimization.
\end{enumerate}

\section*{BACKGROUND}
\subsection * {Kernel Functions}
\textbf{Kernel functions}: Support Vector Machines (SVMs) ~\citep{cortes1995support} utilize a nonlinear transformation $\phi(x)$ to map the input space to a higher-dimensional feature space. This approach employs linear classifiers to tackle nonlinear problems. A variety of kernel functions are utilized, including the linear, polynomial, Gaussian, exponential, and sigmoid kernels\citep{gonen2011multiple}.

\begin{equation}
\text{Linear kernel: } K(\mathbf{x}, \mathbf{y}) = \mathbf{x}^\top \mathbf{y}
\end{equation}
\begin{equation}
\text{Polynomial kernel: } K(\mathbf{x}, \mathbf{y}) = (\gamma \mathbf{x}^\top \mathbf{y} + r)^d
\label{eq:polynomial-kernel}
\end{equation}
\begin{equation}
\text{Gaussian kernel (RBF): } K(\mathbf{x}, \mathbf{y}) = \exp(-\gamma \|\mathbf{x} - \mathbf{y}\|^2)
\label{eq:gaussian-kernel}
\end{equation}
\begin{equation}
\text{Exponential kernel: } K(\mathbf{x}, \mathbf{y}) = \exp(-\gamma \|\mathbf{x} - \mathbf{y}\|)
\label{eq:exponential-kernel}
\end{equation}
\begin{equation}
\text{Sigmoid kernel: } K(\mathbf{x}, \mathbf{y}) = \tanh(\gamma \mathbf{x}^\top \mathbf{y} + r)
\end{equation}
In the above codes, $x$ and $y$ represent input vectors,$\gamma$ and $r$ are parameters that could be specific to the kernel function. The $\exp$ and $\tanh$ functions represent the exponential and sigmoid functions, respectively.

\textbf{Multiple Kernel Learning} (MKL)~\citep{gonen2011multiple,sonnenburg2006largeMKL} is a machine learning method that combines different kernel functions to improve the performance of kernel-based learning algorithms, such as Support Vector Machines (SVM). In MKL, the idea is to learn a linear or non-linear combination of multiple kernels, rather than selecting a single kernel a priori.
The general formulation of the MKL optimization problem for SVM can be written as follows:

\begin{equation}
\min_{w}  \sum_{i=1}^{m} w_i \cdot {K_i(x,y)}
\quad \text{s.t.} w \geq 0, \quad \|w\|_1 = 1
\end{equation}

Let ${K_i(x,y)}$ denote the i-th kernel function, where $w_i$ represents the associated weight with this kernel function. w represents the weight vector containing all the weights associated with the different kernel functions. $\|w\|_1$ denotes the $L1$ norm of $w$, which is the sum of the absolute values of the components in $w$.

\subsection * {Transformer}
The Transformer architecture, introduced by ~\citep{vaswani2017attention,dufter_position_2022}. in their landmark paper "Attention is All You Need", marked a groundbreaking shift in the approach to sequence-to-sequence tasks, abandoning recurrent and convolutional layers in favor of attention mechanisms. The Transformer is an architecture comprising an encoder and a decoder, each composed of multiple stacked layers. Each layer within the encoder incorporates a self-attention mechanism and a feed-forward network. The decoder is similarly organized into stacked layers. We will focus on a particular component of the Transformer known as the attention layer, which is defined as follows: 

\begin{equation}
\text{preSoftmaxA}_{ij} =\sqrt{\frac{1}{d}}q_ik^T_j =\sqrt{\frac{1}{d}}(x_iW^q)(x_jW^k)^T
\label{eq:preSoftmaxA1}
\end{equation}
\begin{equation}
\text{postSoftmaxA}_{ij} = \text{softmax}(\text{preSoftmaxA}_{ij} )
\end{equation}
\begin{equation}
\text{softmax}(\text{preSoftmaxA}_{ij}) = \frac{\exp({\text{preSoftmaxA}_{ij}})}{\sum_{k} \exp({\text{preSoftmaxA}_{ik}})}
\end{equation}
\begin{equation}
{o}_{i} = \text{softmax}(\text{preSoftmaxA})(x_iW^v)^T
\end{equation}
In the equations delineated above, $q_i=x_iW^{q},\ k_i=x_iW^{k},\ and \ v_i=x_iW^{v}$ represent the queries, keys, and values, respectively. These are generated by projecting the input $X$ with the corresponding weight matrices, which are all of dimensions $W^q, \ W^k, \ W^v \in \mathbb{R}^{d \times d}$ with $d$ the hidden dimension. It is common to consider multiple head attention. More specifically, $W^q, \ W^k, \ W^v \in \mathbb{R}^{d \times dh} \ $ where $\ d = hd_h. \ {\text{preSoftmaxA}_{ij}}$ is the score for the j-th key in response to the i-th query.
\subsection*{Length Extrapolation }
When the predicted sequence length surpasses the training length, the inference accuracy of the Transformer model deteriorates\citep{press2021trainalibi}. We endeavor to train the model on a fixed length, L, and guarantee that its performance remains robust when predicting on longer sentences ($L_{test}>L_{train}$). Achieving length generalization necessitates that Transformers effectively generalize to unseen positions during training. The design of improved position encodings constitutes an ongoing research focus aimed at enhancing length generalization.

\subsection*{position encoding}
\textbf{Absolute positional} embeddings assign a positional vector $p_m$ to each position $m$ and add $p_m$ to the embedding vector $e_m$. 

\begin{equation}
\mathbf{E}= [\mathbf{e}_1, \ldots, \mathbf{e}_m, \ldots, \mathbf{e}_n]^\top
\end{equation}
\begin{equation}
\mathbf{P}= [\mathbf{p}_1, \ldots, \mathbf{p}_m, \ldots,\mathbf{p}_n]^\top 
\end{equation}
\begin{equation}
\mathbf{PE}= [\mathbf{e}_1+\mathbf{p}_1, \ldots, \mathbf{e}_m+\mathbf{p}_m,\ldots, \mathbf{e}_n+\mathbf{p}_n]^\top
\end{equation}
Let n represent the length of the sequence, m denotes the m-th position that corresponds to the token, e signifies the embedding of the token, and p denotes the positional encoding. Though simple and straightforward, APE-based Transformers usually generalize poorly to longer
sequences (Press et al., 2022)

\textbf{Relative position} encodings (RPEs)~\citep{su2024roformer,raffel2020exploring,press2021trainalibi,chi2022kerple,li_functional_2023_fire,peng_yarn_2023,jacot_neural_2018_ntk} are a highly regarded form of position encoding in the field. Within natural language processing (NLP) tasks, emphasis is placed on relative position rather than absolute position. It is widely accepted that in natural language the relative position is more significant than absolute position~\citep{raffel2020exploring,chen2021simple}. RPEs utilize relative rather than absolute position information to encode positional context. \cite{neishi2019relation} demonstrate that RPEs exhibit length extrapolation capabilities and greater resilience to length variations. The alignment of RPEs with the characteristics of NLP tasks has led to numerous proposals of RPE methodologies to address length extrapolation challenges. A substantial body of research has been dedicated to RPEs, which can be mathematically represented as follows.

\begin{equation}
\text{preSoftmaxA}_{ij}=q_i k_j^T + \mathbf{B(j-i)} = (x_iW^q)(x_jW^k)^T + \mathbf{B(j-i)}
\label{eq:presoftmaxA}
\end{equation}
Eq.~\eqref{eq:presoftmaxA}, illustrates the computation of RPEs, where $q_i k_j^T$ represents the pre-softmax attention scores from query $i$ to key $j$. In contrast to Eq.~\eqref{eq:preSoftmaxA1}, $\text{B(j-i)}$ is a kernel function used to obtain a bias value, signifying the relative positional information between elements $j$ and $i$. Different ALiBi-type RoPEs employ different approaches to construct the bias term $B$. The softmax function is then applied to the post-softmax attention scores, attention weights $\text{postSoftmaxA}_{ij}$ are computed as 

\begin{equation}
\text{postSoftmaxA}_{ij} = \frac{\exp(q_i^T k_j + \mathbf{B(j-i)})}{\sum_{l=1}^L \exp(q_i^T k_l + \mathbf{B(l-i)})}=\frac{\exp(q_i^T k_j)\exp(\mathbf{B(j-i)})}{\sum_{l=1}^L \exp(q_i^T k_l)\exp(\mathbf{B(l-i)})}
\label{eq:postsoftmaxA}
\end{equation}
The matrix $\exp(\text{B(j-i)})$, representing a kernel function, is applied as a bias to the attention logits after softmax normalization. It imposes a penalty on the post-softmax attention scores corresponding to distant query-key pairs, and this penalty intensifies as their distance increases. While the precise formulation of $B$ varies across studies, its primary role is to modulate post-softmax attention weights based on the distance $|j-i|$. Specifically, when $|j-i|$ is large, $\exp(\text{B(j-i)})$ approaches 0, effectively suppressing the post-softmax attention weight; conversely, when $|j-i|$ is small or zero, $\exp(\text{B(j-i)})$ approaches 1, thereby enhancing the attention weight. In this manner, $\exp(\text{B(j-i)})$ acts as a kernel function: significant attention weights are assigned to nearby positions $i$ and $j$, while distant positions are assigned minimal weights.

\textbf{ALiBi} adopts a simpler method to represent relative position information. Here, the scalar m is a head-specific slope that is fixed before training. It is worth noting that the absence of additional learnable parameters contributes to the superior efficiency and potentially enhances the extrapolation capabilities of ALiBi. This could be perceived as a disadvantage when compared to state-of-the-art (SOTA) models.

\begin{equation}
\text{preSoftmaxA}_{ij}=q_i^T k_j -m|j-i|
\end{equation}
\begin{equation}
\exp(\mathbf{B(j-i)}) = \exp(\mathbf{-m(j - i)}) = \exp(-m|j-i|) = \frac{1}{\exp(m|j-i|)}
\label{eq:presoftmaxA-alibi}
\end{equation}
Eq.~\eqref{eq:presoftmaxA-alibi} is Exponential kernel. For instance, with a sentence length of 512 and $j-i=512$, the softmax value becomes negligible and approaches zero. Conversely, when m-n=0, the softmax value equals 1.

\textbf{Kerple} introduces the concept of kernel functions through the use of kernelized positional differences and proposes two types of bias kernel functions: polynomial and Gaussian-like. They formulate the equation:

\begin{equation}
\text{preSoftmaxA}_{ij} = q_i^T k_j - r_1 \cdot \log(1 + r_2 |j-i|)
\end{equation}
\begin{equation}
\exp(\mathbf{B(j-i)}) = \exp(-r_1 \log(1 + r_2 |j-i|)) = \frac{1}{(1 + r_2 |j-i|)^{r_1}}
\label{eq:presoftmaxA-kerple}
\end{equation}

where $r_1$ and $r_2$ are positive scalar parameters, each specific to a given layer. Eq.~\eqref{eq:presoftmaxA-kerple} is Polynomial kernel. In Kerple, the kernel function $\exp(\text{B(j-i)})$ is characterized as same as Eq.~\eqref{eq:presoftmaxA-alibi}: 
This relationship is consistent with the principle that a large $|j-i|$ corresponds to a reduced attention weight, whereas a small $|j-i|$ corresponds to an increased attention weight. The paper~\citep{chi2022kerple} observes that ALiBi and its generalized power variant can quickly reach highly negative values. In contrast, the log variant successfully discerns several flat kernels, thereby effectively extending the window of attention.

\textbf{Sandiwish} represents a pioneering parameter-free approach in relative positional embedding, relying exclusively on the inner product of two position embeddings. This approach retains only the inner product of two position embeddings. Notably, in this formalization,  $p_i^Tp_j$ becomes the temporal bias term, exhibiting the same decay-with-distance pattern as ALiBi, aligning precisely with the authors' objectives. 

\begin{equation}
\text{preSoftmaxA}_{ij}=q_i^T k_j + p_i^T p_j
\end{equation}
\begin{equation}
\exp(\mathbf{B(j-i)}) =\exp(p_i^Tp_j)
\label{eq:presoftmaxA-sandiwish}
\end{equation}

Eq.~\eqref{eq:presoftmaxA-sandiwish} is Exponential kernel. The Sandiwish kernel function $\exp(\text{B(j-i)})$ is defined as  
As $\text{i-j}$ approaches zero, the $(p_ip_j)$ similarity markedly increases. When  $(p_ip_j)$ this value equals one, the exponential function attains its maximum. Conversely, when the $(p_ip_j)$ similarity equals zero, the exponential function reaches its minimum.

\textbf{FIRE} formula for B utilizes a learnable continuous function, enabling it to address inputs beyond the training domain through progressive interpolation, and can encode positions for sequences of any length within the 0 to 1 range.

\textbf{T5}, the attention score $q_i^Tk_j$ is defined as 

\begin{equation}
q_i^T k_j = (x_i W^q)(x_j W^k)^T + b(i, j)
\end{equation}

The bias term $b(i,j)$ is given by $b(i, j) = r_{\min\{|i-j|, K\}}$, where $K$ is a hyper-parameter and ${\{r_i\}}_{i=0}^{K}$ are learnable scalars. T5 employs a logarithmic bucket assignment which effectively allows it to extrapolate across various input lengths.

In summary, all the discussed mechanisms employ kernel functions with the goal of decreasing the post-softmax attention weight as the difference between positions j and i increases, and conversely, increasing it as the difference decreases. However, the aforementioned B functions in relative position encoding utilize single kernel functions and do not leverage the benefits of multiple kernels. While mechanisms like Kerple, FIRE, and T5 are effective, they introduce additional parameters and computational complexity, thus presenting a trade-off between performance and efficiency. On the other hand, ALiBi and Sandiwish do not depend on additional parameters, yet they tend to yield weaker results.
\section*{Methods}
In this section, we detail the MEP method and how to use multiple kernel learning to construct the bias term. Our method is simple and can be seamlessly integrated with any ALiBi-type positional encoding method. This method represents a novel relative positional encoding approach that employs Multiple Kernel Learning to obtain the bias term, which penalizes the post-softmax attention scores. It synergizes the advantages of different kinds of kernels, such as the exponential kernel and Gaussian kernel, to address the limitations inherent in traditional approaches like T5, ALiBi, Kerple, and Sandwiched, which use a single kernel function for positional encoding and do not fully exploit the advantages of each kernel function. This method has been shown to enhance length extrapolation performance. See schematic diagram \ref{fig:alibi-mep}

\begin{equation}
\text{Exponential Kernel}=\exp(\mathbf{B(j-i)})= \exp(-r_1|j-i|) = \frac{1}{\exp(r_1|j-i|)}
\label{eq:expB-exponential}
\end{equation}
\begin{equation}
\text{Gaussian Kernel}= \exp(\mathbf{B(j-i)})=\exp(-r_2 \|{j} -{i}\|^2)=\frac{1}{\exp(r_2 \|{j} -{i}\|^2)}
\label{eq:expB-gaussian}
\end{equation}
\begin{equation}
\text{our-parameter-free}((\mathbf{B(j-i)})) = \log(\alpha \frac{1}{\exp(r_1|j-i|)} + \beta \frac{1}{\exp(r_3|j-i|)}+\gamma\frac{1}{\exp(r_2\|{j} -{i}\|^2)})
\label{eq:presoftmax-our-nofree}
\end{equation}
\begin{equation}
\text{our-parameter-free Kernel} = \exp(\mathbf{B(j-i)}) = \alpha \cdot \text{exponential} + \beta \cdot \text{exponential} + \gamma \cdot \text{Gaussian}
\label{eq:postsoftmax-our-nofree}
\end{equation}
\begin{equation}
\text{our-parameter}((\mathbf{B(j-i)})) = \log(\alpha \frac{1}{(1 + r_2 |j-i|)^{r_1}} + \beta\frac{1}{\exp(r_3 \|{j} -{i}\|^2)})
\label{eq:presoftmax-our-para}
\end{equation}
\begin{equation}
\text{our-parameter Kernel} = \exp(\mathbf{B(j-i)}) = \alpha \cdot \text{Kerple-log kernel} + \beta \cdot \text{Gaussian}
\label{eq:postsoftmax-our-para}
\end{equation}

Eq.~\eqref{eq:expB-exponential} is the same as Eq.~\eqref{eq:exponential-kernel}. Eq.~\eqref{eq:expB-gaussian} is the same as Eq.~\eqref{eq:gaussian-kernel}. While Eq.~\eqref{eq:presoftmax-our-nofree} represents our MEP method applied before the softmax operation (More details see Eq.~\eqref{eq:presoftmaxA}) in the parameter-free model. Eq.~\eqref{eq:postsoftmax-our-nofree} represents the bias term after the softmax operation. In our MEP method, it merges the exponential and Gaussian kernels. Eq.~\eqref{eq:presoftmax-our-para} and Eq.~\eqref{eq:postsoftmax-our-para} are the same as Eq.~\eqref{eq:presoftmax-our-nofree} and Eq.~\eqref{eq:postsoftmax-our-nofree}, respectively. They represent the parameter model. This model fuses the Kerple-log and Gaussian kernels. $r_1$, $r_2$, and $r_3$ are all slope values. For more information, refer to the ~\citep{press2021trainalibi} paper. $r_1$ and $r_2$ are equal to the ALiBi slope values, while $r_3$ slope value is equal to the ALiBi slope values multiplied by 0.5.

\textbf{Determining which kernel functions to select and how to combine multiple kernel functions?} After conducting numerous ablation experiments and a theoretical demonstration, we ultimately opted for Eq.~\eqref{eq:expB-exponential} the exponential and Eq.~\eqref{eq:expB-gaussian} Gaussian kernel functions in the parameter-free approach, we chose the Gaussian and Kerple-log kernel functions in the parameterized approach, see in Eq.~\eqref{eq:postsoftmax-our-nofree} and Eq.~\eqref{eq:postsoftmax-our-para}. 

\cite{chi2022kerple} mentioned that both ALiBi and its generalized power variant rapidly assume highly negative values. In contrast, the log variant has been shown to identify several flat kernels, effectively extending the range of post-softmax attention scores. Consequently, we opt for a function with a flatter profile that is close to zero. The Multiple Kernel Learning (MKL) methodology is utilized, in conjunction with a summation strategy, to ensure that the final expression for the $\exp(B)$ bias term slowly approaches zero. We employ a straightforward weighted average approach to combine individual kernel functions. In the MKL parameter-free approach configuration, We selected $\exp(x)$, $\text{Gaussian}(x)$, and $\exp(0.5*x)$ as kernel functions, where $\exp(0.5*x)$ represents an exponential kernel at half the slopes value of ALiBi, setting $\alpha=0.33$, $\beta=0.33$, and $\gamma=0.33$. In the MKL parameterized approach configuration, we chose $\text{Gaussian}(x)$ and Kerple-log kernels, setting $\alpha=0.5$ and $\beta=0.5$. Additional kernel functions and combination methods may also be explored.

Concurrently, our experimental design will entail a comparative analysis of various kernel functions' effects. This includes assessing the exponential, polynomial, and Gaussian kernels, comparing the merits and drawbacks of assorted kernel function combinations, as well as evaluating the strengths of individual kernel functions.

\textbf{Criteria for Selecting Slopes for Individual Kernel Functions.}
The selection of slopes plays a crucial role in the performance of kernel functions~\citep{chi_dissecting_2023}. Each kernel function requires specific slope values to achieve optimal results. For the exponential kernel function, the slopes are derived from ALiBi. The slopes for the Gaussian kernel function are similar to those of the exponential kernel function. In the parameterized approach, the Kerple-log kernel utilizes a learnable approach to optimize the slopes. For further details, refer to the related ablation study presented in table \ref{tab:slopes-and-8t2}.

\textbf{Add bias to post-softmax attention scores}
The final bias term is computed using cumulative summation, where each bias is the product of a weighted average coefficient and the each kernel function result (the kernel inputs include the slope value and the i-j distance). see Eq.~\eqref{eq:postsoftmaxA} ~\eqref{eq:postsoftmax-our-nofree} ~\eqref{eq:postsoftmax-our-para}. This bias is seamlessly incorporated into the post-softmax attention scores, thereby effectively modulating penalties at varying rates.

\section*{experiment}
\subsection*{Dataset and Implementation Description}
\paragraph{Dataset.} To evaluate the performance of our algorithm, we conducted experiments using the OpenWebText2, GitHub, and ArXiv datasets\citep{gao2020pile}. OpenWebText2 is a large, filtered corpus of text documents obtained from URLs found in Reddit submissions. The plug-and-play version of OpenWebText2 contains 17,103,059 documents, totaling 65.86GB of uncompressed text. GitHub is an AI-powered development platform that enables developers to create, store, and manage code, supporting various programming languages such as C/C++, Java, and Python, among others. ArXiv is a repository of scientific papers spanning fields such as mathematics, physics, astronomy, electrical engineering, computer science, quantitative biology, statistics, mathematical finance, and economics.
\paragraph{Implementation.} Our experimental configuration is similar to that of the Kerple paper, see table~\ref{tab:model_configs}; we have adapted our model from GPT-NeoX~\citep{gpt-neox}, which is a Transformer model implemented by the EleutherAI team. The codebase is based on the NVIDIA Megatron Language Model~\citep{shoeybi2019megatron} and is further optimized using the Microsoft DeepSpeed library~\citep{rasley2020deepspeed}. Our model was trained on a system equipped with a single NVIDIA A100 GPU with 40 GB of memory. We retained most settings from the small GPT-NeoX configuration, except for changing the train-micro-batch-size to 16, gradient-accumulation-steps to 2, and setting the attention-softmax-in-fp32 flag to True, the train sequence length to 512, and the maximum position embeddings to 512.
\subsection*{Experimental Results}
In this chapter, we conduct an experimental comparison of the proposed MEP, which is based on the MKL relative position encoding ALiBi-type method, with the open-source ALiBi, T5, and Kerple methods. The experimental data consist of the OpenWebText2, GitHub, and arXiv datasets. The experimental results demonstrate that our proposed method (be it parameter-free or parameter-based) exhibits significant improvements in length extrapolation when the training length is smaller than the predicted length. We first compare the perplexity values on inputs with different lengths (512 to 8192) from various datasets to evaluate the long-context generalization ability of different position encoding methods. Second, We present several ablation experiments to investigate the design choices of MEP. We assess the impact of various combinations of kernel functions by conducting an ablation study on the GitHub and arXiv datasets. We also analyze how varying the slopes of ALiBi influences the final perplexity and conduct ablation studies on these variations. Finally, we present training time comparisons across various methods. In the discussion section, we elucidate why the MKL approach surpasses existing methods and demonstrate this with illustrative figures.
\paragraph{main Experimentanl Results.} From the data presented in the table \ref{tab:openweb-github-arxiv}, we confirmed the effectiveness of our method across three datasets. We benchmarked our method against mainstream open-source ALiBi-type models, namely ALiBi, T5, and Kerple. The left column of the table lists the non-parametric models, while the right column provides details of the models with trainable parameters. Overall, both non-parametric and parametric models attained lower perplexity scores across the three datasets. 

On the non-parametric models, our method consistently outperforms the ALiBi method, except for the length of 512 on openwebtext2 dataset, where it is slightly worse than ALiBi by 0.05 points. 

On the parametric models, by simply choosing the ALiBi and Kerple-log polynomial kernel functions via the MKL method and employing the average coefficient approach to combine these functions, our method exceeded the performance of the state-of-the-art Kerple method on openwebtext2 (in four out of five cases) and github dataset (in four out of five cases), but  did not exceed Kerple on the arXiv dataset. For sequence lengths of 4096 and 8192, the MEP model outperforms T5 across all datasets. However, at the sequence lengths of 512 , 1024 and 2048, the T5 method outperformed our approach. The integration of the T5 method will be considered in future research endeavors.

From an overall perspective, regardless of the non-parametric or parametric fusion method, our method has surpassed the individual kernel functions, this demonstrates that our method has a significant impact on length extrapolation. However, at the sequence lengths of 512 , 1024 and 2048, the T5 method outperformed our approach. The integration of the T5 method will be considered in future research endeavors.

\begin{table*}[!ht]
\centering
\setlength{\tabcolsep}{2pt}
\hspace{-3.5mm}
\begin{tabular}{@{\extracolsep{3pt}}lcccccc}
\hline\hline
\multicolumn{7}{c}{\textbf{OpenWebText2}}\\
\midrule
{Length} & ours & ALiBi & & {ours-parameters} &{KERPLE} & {T5}\\
\midrule
512  & 23.90 $\pm$ 0.54 & 23.85 $\pm$ 0.37 & & 23.77 $\pm$ 0.47 & 23.88 $\pm$ 0.38 & \textbf{23.71 $\pm$ 0.36} \\
1024 & \textbf{21.92 $\pm$ 0.40} & 22.14 $\pm$ 0.41 & & \textbf{21.75 $\pm$ 0.32} & 22.02 $\pm$ 0.46 & 21.90 $\pm$ 0.39 \\
2048 & 21.99 $\pm$ 0.01 & 21.99 $\pm$ 0.03 & & 21.68 $\pm$ 0.02 & \textbf{21.66 $\pm$ 0.05} & 21.76 $\pm$ 0.05 \\
4096 & \textbf{21.55 $\pm$ 0.31} & 21.65 $\pm$ 0.17 & & \textbf{21.23 $\pm$ 0.23} & 21.27 $\pm$ 0.19 & 22.40 $\pm$ 0.14 \\
8192 & \textbf{21.57 $\pm$ 0.13} & 21.72 $\pm$ 0.09 & & \textbf{21.27 $\pm$ 0.15} & 21.30 $\pm$ 0.14 & 25.93 $\pm$ 0.84 \\
\hline\hline
\multicolumn{7}{c}{\textbf{GitHub}}\\
\midrule
{Length} & ours & ALiBi &{} & {ours-parameters} &{KERPLE} & {T5}\\
\midrule
512  & 2.607 $\pm$ 0.015 & 2.607 $\pm$ 0.014 & & 2.604 $\pm$ 0.015 & 2.604 $\pm$ 0.015 & \textbf{2.588 $\pm$ 0.015} \\
1024 & \textbf{2.436 $\pm$ 0.003} & 2.450 $\pm$ 0.003 & & 2.422 $\pm$ 0.003 & 2.421 $\pm$ 0.003 & \textbf{2.406 $\pm$ 0.003} \\
2048 & \textbf{2.394 $\pm$ 0.005} & 2.430 $\pm$ 0.006 & & 2.325 $\pm$ 0.004 & 2.325 $\pm$ 0.005 & \textbf{2.323 $\pm$ 0.004} \\
4096 & \textbf{2.345 $\pm$ 0.009} & 2.386 $\pm$ 0.008 & & \textbf{2.239 $\pm$ 0.007} & 2.242 $\pm$ 0.006 & 2.315 $\pm$ 0.010 \\
8192 & \textbf{2.318 $\pm$ 0.004} & 2.360 $\pm$ 0.004 & & \textbf{2.211 $\pm$ 0.002} & 2.221 $\pm$ 0.003 & 2.531 $\pm$ 0.008 \\
\hline\hline
\multicolumn{7}{c}{\textbf{ArXiv}}\\
\midrule
{Length} & ours & ALiBi &{} & {ours-parameters} &{KERPLE} & {T5}\\
\midrule
512 & \textbf{6.316 $\pm$ 0.014} & 6.324 $\pm$ 0.014 && 6.303 $\pm$ 0.015 & 6.302 $\pm$ 0.013 & \textbf{6.254 $\pm$ 0.012} \\
1024 & \textbf{5.612 $\pm$ 0.039} & 5.640 $\pm$ 0.042 && 5.576 $\pm$ 0.040 & 5.570 $\pm$ 0.039 & \textbf{5.528 $\pm$ 0.038} \\
2048 & \textbf{5.511 $\pm$ 0.039} & 5.611 $\pm$ 0.036 && 5.332 $\pm$ 0.036 & 5.328 $\pm$ 0.036 & \textbf{5.322 $\pm$ 0.032} \\
4096 & \textbf{5.230 $\pm$ 0.004} & 5.356 $\pm$ 0.003 && 4.902 $\pm$ 0.002 & \textbf{4.895 $\pm$ 0.005} & 5.072 $\pm$ 0.006 \\
8192 & \textbf{5.309 $\pm$ 0.013} & 5.440 $\pm$ 0.015 && 4.936 $\pm$ 0.016 & \textbf{4.914 $\pm$ 0.021} & 5.907 $\pm$ 0.885 \\
\hline\hline
\end{tabular}
\caption{\textbf{Perplexity Comparison on the OpenWebText2, GitHub, and ArXiv datasets.} All models are trained for 50k steps with a training length of 512 and five random seeds. The models listed in the left section feature parameter-free positional embeddings; "ours-non-parameters" indicates the absence of learnable parameters. In contrast, both KERPLE and T5 possess positional embeddings with learnable parameters; "ours-parameters" denotes the presence of learnable parameters.}
\label{tab:openweb-github-arxiv}
\end{table*}

\paragraph{ablation for kernel functions.}To assess the impact of our proposed method, we conducted a series of ablation studies that focused on various kernel function combinations. The experimental results indicate that different combinations yield distinct levels of effectiveness. As illustrated in the accompanying table \ref{tab:ablation-openweb-github-arxiv}, on both the OpenWebText2 and GitHub datasets, the application of the Multiple Kernel Learning (MKL) technique improved the performance of both non-parametric and parametric methods compared to their states prior to fusion.

For the non-parametric model, performance improvement was observed with the merging of either two or three kernel functions. Notably, the combination of three kernel functions across both datasets resulted in a more favorable perplexity outcome compared to the original singular ALiBi kernel approach, with the exception of the sequence length of 512. For instance, within the OpenWebText2 dataset, merging the ALiBi and scaled ALiBi (ALiBi*0.5) kernel functions demonstrated improved extrapolation capabilities over the baseline ALiBi method at sequence lengths of 2048, 4096, and 8192. Similarly, the fusion of ALiBi and Gaussian kernel functions outperformed the standalone ALiBi method at sequence lengths of 512 and 1024. It is hypothesized that the divergent kernel functions prioritize different aspects of post-softmax attention, which contributes to the observed variations in performance.

In the parametric model, it was observed that the combination of Kerple and Gaussian kernel functions generally resulted in enhanced performance (in four out of five cases). Furthermore, the integration of multiple kernel functions significantly improved the model's ability to extrapolate effectively at an input length of 8192(except for Kerple-log + ALiBi), while preserving performance at other sequence lengths.

Despite these advancements, the performance at a sequence length of 512 has yet to surpass that of the T5 method. Nonetheless, our method did outperform the Kerple-log approach. This finding underscores the necessity for continued optimization and enhancement efforts.

\begin{table*}[!ht]
\centering
\setlength{\tabcolsep}{2pt}
\hspace{-3.5mm}
\begin{tabular}{@{\extracolsep{3pt}}lcccc}
\hline\hline
\multicolumn{5}{c}{\textbf{OpenWebText2-non-parametri}}\\
\midrule
{Length} & ALiBi & {ALiBi+ALiBi*0.5} & {ALiBi+Gaussian} &{ALiBi+ALiBi*0.5+Gaussian}\\
\midrule
512  & 23.85 $\pm$ 0.37 & 23.94 $\pm$ 0.41 & \textbf{23.78 $\pm$ 0.49} & 23.90 $\pm$ 0.54 \\
1024 & 22.14 $\pm$ 0.41 & 22.16 $\pm$ 0.45 & \textbf{21.88 $\pm$ 0.36} & 21.92 $\pm$ 0.40 \\
2048 & 21.99 $\pm$ 0.03 & \textbf{21.97 $\pm$ 0.02} & 22.04 $\pm$ 0.01 & 21.99 $\pm$ 0.01 \\
4096 & 21.65 $\pm$ 0.17 & 21.58 $\pm$ 0.25 & 21.65 $\pm$ 0.27 & \textbf{21.55 $\pm$ 0.31} \\
8192 & 21.72 $\pm$ 0.09 & 21.62 $\pm$ 0.12 & 21.73 $\pm$ 0.05 & \textbf{21.57 $\pm$ 0.13} \\
\hline\hline
\multicolumn{5}{c}{\textbf{OpenWebText2-parametri}}\\
\midrule
{Length} & Kerplg-log & {Kerplg-log+ALiBi} & {Kerplg-log+Gaussian} &{Kerplg-log+ALiBi+Gaussian}\\
\midrule
512  & 23.88 $\pm$ 0.38 & 23.82 $\pm$ 0.46 & \textbf{23.77 $\pm$ 0.47} & 23.84 $\pm$ 0.48 \\
1024 & 22.02 $\pm$ 0.46 & 21.83 $\pm$ 0.38 & \textbf{21.75 $\pm$ 0.32} & 21.80 $\pm$ 0.33 \\
2048 & \textbf{21.66 $\pm$ 0.05} & 21.75 $\pm$ 0.03 & \textbf{21.68 $\pm$ 0.02} & 21.71 $\pm$ 0.04 \\
4096 & 21.27 $\pm$ 0.19 & 21.31 $\pm$ 0.33 & \textbf{21.23 $\pm$ 0.23} & 21.24 $\pm$ 0.32 \\
8192 & 21.30 $\pm$ 0.14 & 21.31 $\pm$ 0.15 & 21.27 $\pm$ 0.15 & \textbf{21.22 $\pm$ 0.19} \\
\hline\hline

\multicolumn{5}{c}{\textbf{GitHub-non-parametri}}\\
\midrule
{Length} & ALiBi & {ALiBi+ALiBi*0.5} & {ALiBi+Gaussian} &{ALiBi+ALiBi*0.5+Gaussian}\\
\midrule
512  & 2.607 $\pm$ 0.014 & \textbf{2.605 $\pm$ 0.015} & 2.613 $\pm$ 0.015 & 2.607 $\pm$ 0.015 \\
1024 & 2.450 $\pm$ 0.003 & \textbf{2.434 $\pm$ 0.003} & 2.451 $\pm$ 0.003 & 2.436 $\pm$ 0.003 \\
2048 & 2.430 $\pm$ 0.006 & \textbf{2.388 $\pm$ 0.005} & 2.429 $\pm$ 0.007 & 2.394 $\pm$ 0.005 \\
4096 & 2.386 $\pm$ 0.008 & \textbf{2.340 $\pm$ 0.007} & 2.386 $\pm$ 0.008 & 2.345 $\pm$ 0.009 \\
8192 & 2.360 $\pm$ 0.004 & \textbf{2.293 $\pm$ 0.003} & 2.361 $\pm$ 0.006 & 2.318 $\pm$ 0.004 \\
\hline\hline

\multicolumn{5}{c}{\textbf{GitHub-parametri}}\\
\midrule
{Length} & Kerplg-log & {Kerplg-log+ALiBi} & {Kerplg-log+Gaussian} &{Kerplg-log+ALiBi+Gaussian}\\
\midrule
512  & \textbf{2.604 $\pm$ 0.015} & 2.606 $\pm$ 0.014 & \textbf{2.604 $\pm$ 0.015} & 2.606 $\pm$ 0.015 \\
1024 & \textbf{2.421 $\pm$ 0.003} & 2.423 $\pm$ 0.003 & 2.422 $\pm$ 0.003 & 2.423 $\pm$ 0.003 \\
2048 & \textbf{2.325 $\pm$ 0.005} & 2.327 $\pm$ 0.004 & \textbf{2.325 $\pm$ 0.004} & 2.326 $\pm$ 0.004 \\
4096 & 2.242 $\pm$ 0.006 & 2.245 $\pm$ 0.005 & 2.239 $\pm$ 0.007 & \textbf{2.238 $\pm$ 0.006} \\
8192 & 2.221 $\pm$ 0.003 & 2.214 $\pm$ 0.005 & 2.211 $\pm$ 0.002 & \textbf{2.206 $\pm$ 0.003} \\
\hline\hline
\end{tabular}
\caption{\textbf{Perplexity Comparison on the OpenWebText2, GitHub datasets.} Ablation studies were conducted to focus on various kernel function combinations using the OpenWebText2 and GitHub datasets. Non-parametric models are parameter-free, whereas parametric models are characterized by learnable parameters. In these studies, $ \text{ALiBi*0.5} = \exp(\text{-0.5} \times \text{x})$, representing an exponential decay at half the rate of ALiBi, and Gaussian refers to a Gaussian kernel function. Furthermore, the '+' symbol indicates the use of Multiple Kernel Learning (MKL) for merging kernel functions.}
\label{tab:ablation-openweb-github-arxiv}
\end{table*}

\paragraph{ablation for alibi slopes.}
The importance of slopes cannot be overstated. Different head slope values impose penalties at varying rates depending on the slope magnitude, which is crucial for length extrapolation. This section investigates the influence of various slope values on the extrapolative effect. As detailed in the [ref] paper, uniform slopes were used for each head, differing from the original geometric sequence by setting all slopes to a uniform value. The slope value was then incrementally increased from 2 to 8 in steps of 2. Upon closer examination of the impact of varying slope values, it was observed that as the head number increased, the slope value decreased, resulting in a decline in the extrapolation effect. For head=6, head=8, and head=9, the extrapolation effect was poor, almost non-existent.

For head=6, head=8, and head=9, adverse effects associated with their slope values have been observed. Consequently, by removing the unfavorable slope values associated with each head and substituting them with beneficial ones, one may inquire whether this would result in enhanced length extrapolation? This is exemplified by transformations such as 8t2, which involves altering the slope value associated with head=8 to slope value of head=2, and similar adjustments like 8t4, 6t2, 6t4, 8t9, 6t9, among others. Table \ref{tab:slopes-and-8t2} reveals that 8t2 or 8t4 does not alter the effect(worse results in four out of five cases); conversely, replacing 6t9(better results in three out of five cases) or 8t9 (better results in one out of five cases) improves the outcome. 

\begin{figure}
    \centering
    \includegraphics[width=\linewidth]{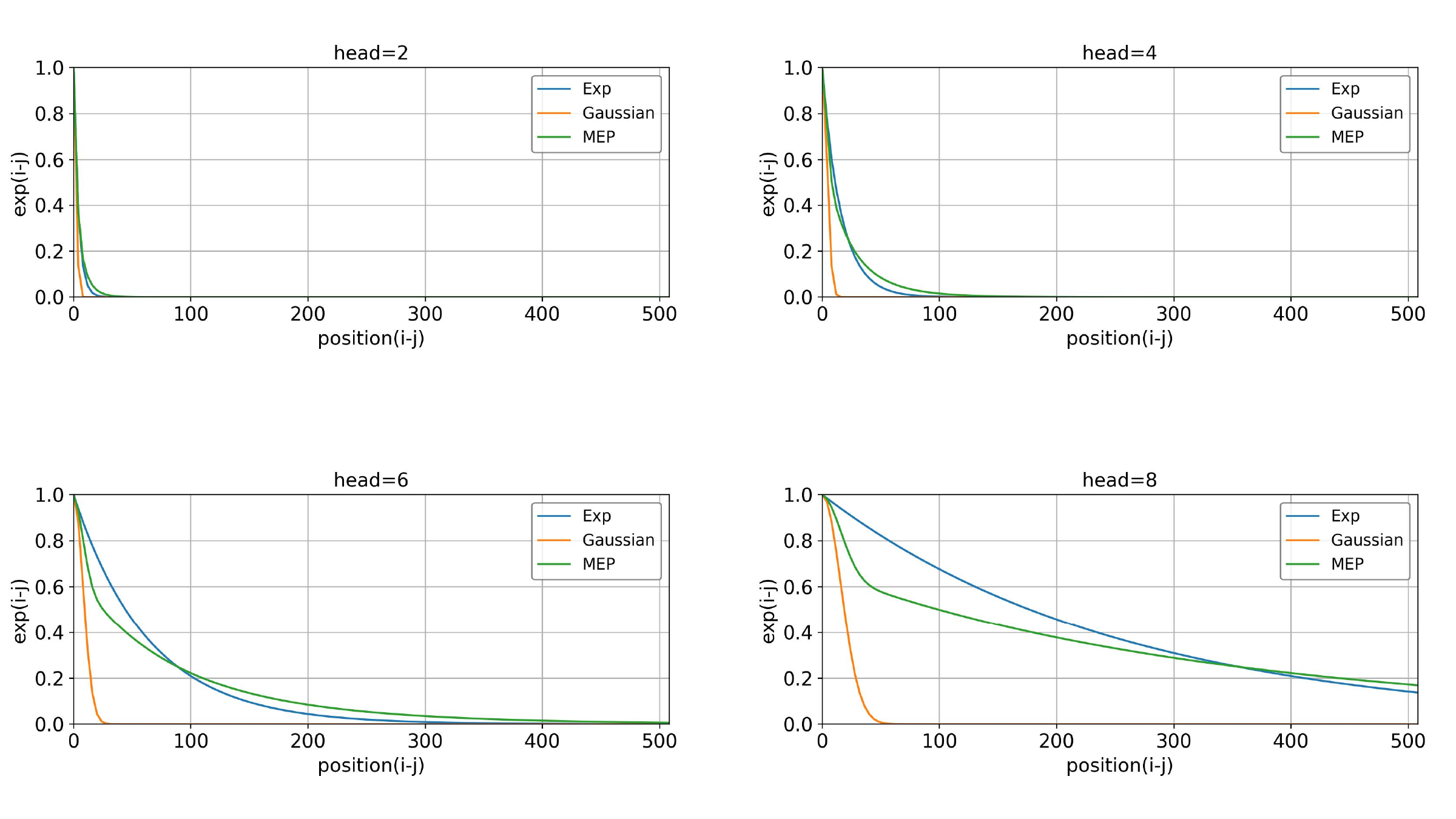}
    \caption{Exponential, Gaussian, and MEP function curves. The x-axis represents the relative position, $i-j$, from 0 to 511; the y-axis represents the value after the kernel function is applied.} 
    \label{fig:mep-lineplot}
\end{figure}

As illustrated in Figure \ref{fig:mep-lineplot}, with slope values corresponding to head = 2 or head = 4, exp(x), for distances greater than 10 ($|i-j| > 10$), the post-softmax attention scores rapidly approach zero. This results in a diminished capacity to attend to more extended distances. However, for head = 6 or head = 8, when considering distances equal to the training length of $|i-j|=512$, the post-softmax attention scores do not equal zero, thereby facilitating increased attention to longer distances and consequently enhancing extrapolation capability.

\begin{table*}[!ht]
\centering
\setlength{\tabcolsep}{2pt}
\hspace{-3.5mm}
\begin{tabular}{@{\extracolsep{3pt}}lccccccccccccc}
\hline\hline
\multirow{2}{*}{Length} & \multicolumn{6}{c}{Same $h$ for all heads} & & \multicolumn{6}{c}{Replace head slopes value} \\
\cline{2-7} \cline{9-14} &
$h$:default & 2 & 4 & 6 & 8 & 9 & & 8t2 & 8t4 & 6t2 & 6t4 & 8t9 & 6t9 \\
\hline
512 & 23.9 & 27.9 & 24.9 & 25.1 & 34.7 & 25.7 &  & 23.9 & 23.9 & 23.9 & 23.9 & 23.9 & 23.9 \\
1024 & 22.1 & 26.2 & 25.2 & 43.6 & 54.8 & 64 &  & 22.2 & 22.3 & 22.1 & 22.2 & 22.1 & 22.1 \\
2048 & 22.0 & 26.5 & 27.2 & 88.8 & 274.4 & 1406 &  & 22.3 & 22.4 & 22.0 & 22.0 & 22.0 & \textbf{21.9} \\
4096 & 21.7 & 26.1 & 27.6 & 124.7 & 938.2 & 13337 &  & 22.0 & 22.2 & \textbf{21.6} & \textbf{21.6} & \textbf{21.6} & \textbf{21.5} \\
8192 & 21.7 & 26.3 & 28.6 & 150.2 & 1795 & 50282 &  & 22.2 & 22.3 & \textbf{21.6} & 21.7 & 21.7 & \textbf{21.5} \\
    \hline\hline
    \end{tabular}
    \caption{The two experiments on the openWebText2 dataset. On the left, the perplexity is shown when a same slope value is used for all heads. On the right, the perplexity is presented after replacing the individual head's slope value.}
    \label{tab:slopes-and-8t2}
\end{table*}

\paragraph{Visualization} In Figure \ref{fig:alibi-gaussian-postsoftmax-attentions}, each point denotes the post-softmax attention score corresponding to the relative position $|i-j|$. The figure presents diagrams of post-softmax attention scores for multiple kernel functions across a range of heads, each capturing attention over varying distances; for instance, when head=2, the attention is focused on shorter distances, whereas head=8 extends to longer distances. In these diagrams, dark blue indicates a score equal to 1, while white signifies a score of 0. Points located in the area above the diagonal line are excluded from consideration. For ease of visualization, scores have been magnified by a factor of 100.

In Figure \ref{fig:alibi-gaussian-postsoftmax-attentions}(a), as described by the formula (\ref{eq:presoftmaxA-alibi}), when the head is set to 2, the blue area is confined to a narrow region, indicating that only a subset of the relative positions $|i-j|$ receives a non-zero post-softmax attention score. As the value of the head increases, for instance, head to 8, the extent of the blue region expands, indicating that attention scores are non-zero, suggesting that the model is capable of focusing on more extended distances.
Similarly, the attention scores derived from the Gaussian kernel function in Figure \ref{fig:alibi-gaussian-postsoftmax-attentions}(b) and the MEP method in Figure \ref{fig:alibi-gaussian-postsoftmax-attentions}(c) exhibit analogous trends.

Upon comparing Figures \ref{fig:alibi-gaussian-postsoftmax-attentions} (a), (b), and (c), and in conjunction with Figure \ref{fig:mep-lineplot}, which illustrates the different ALiBi, Gaussian, and MEP models, it is observable that the MEP model's scores approach zero more gradually and smoothly, whereas the ALiBi model's scores diminish more rapidly, and the Gaussian model's scores decline the most precipitously. For instance, with head set to 6, in the lower left corner, ALiBi's score at position [511, 0] is 0.00035, Gaussian's is 0, and MEP's is 0.0062.

\paragraph{Training Time}
In Table \ref{tab:train-speed}, we can see that our parameter-free model has a training time that is equal to the ALiBi time. Our parameterized model has a training time that is equal to or greater than the Kerple-log time, but both are less than the T5's time.

\begin{table}[!ht]
    \centering
    \caption{\textbf{Training Time Comparison on GitHub/OpenWebText2/ArXiv}}
    \begin{tabular}{lccccc}
    \hline\hline
    sec/step & {ALiBi} &{T5} & {KERPLE} & {ours-parameter-free} &{ours-parameters}\\
    \hline
    Github & 0.36 & 0.42 & 0.39 & 0.36 & 0.38 \\
    Openwebtext2 & 0.36 & 0.42 & 0.38 & 0.36 & 0.39 \\
    ArXiv & 0.36 & 0.42 & 0.37 & 0.36 & 0.39 \\
    \hline\hline
    \end{tabular}
    \label{tab:train-speed}
\end{table}

\section*{Discussion}
This paper has presented both theoretical and experimental analyses, demonstrating that the proposed MEP for Relative Positional Encoding method is valid. It based on Multi-Kernel Learning (MKL)~\citep{gonen2011multiple,sonnenburg2006largeMKL}, effectively fuses multiple kernel functions to construct biases that influence post-softmax attention scores, surpassing existing non-parametric and parametric methods.

Our literature review identified recent developments in positional encoding (PE) have revealed that existing alibi-type relative position coding exhibits limitations~\citep{chi2022kerple,chi_dissecting_2023,li_functional_2023_fire,zhao2023length}. With regard to bias generation, these methods employ a singular kernel function, which fails to leverage the full potential of kernel-based approaches and lacks a mechanism to integrate the benefits of both established and emerging methods. To date, there has been no exploration of integrating multiple methods to achieve bias optimization, and this remains an uncharted area of research.

Our approach comprises two methodologies: non-parametric (eg. $\exp+\text{gaussian}+\exp(0.5)$) and parametric (eg. $\text{kerple}+\text{gaussian}$). Upon fusing the kernel functions, we derive the bias, which is then integrated with the head slopes to influence the post-softmax attention scores. This integrated bias positional encoding achieves state-of-the-art (SOTA) performance on three datasets, highlighting the efficacy of our kernel function fusion method. We conducted ablation studies with diverse kernel function combinations, and the results demonstrated that each combination yielded varying degrees of effectiveness. Furthermore, we assessed the impact of varying slope values on the outcomes. Our findings demonstrate that varying head configurations and slope values cause the model to focus on different sequence lengths, producing divergent effects. These fundamental observations align with research~\citep{chi_dissecting_2023} indicating that slope values impose penalties at differential rates based on their magnitude, a key factor in sequence length extrapolation. The aforementioned results validate the effectiveness of our proposed MEP method.

~\cite{chi2022kerple} suggests that smoothing the bias correlates with enhanced model performance. Analysis of post-softmax attention scores heatmaps(figure \ref{fig:alibi-gaussian-postsoftmax-attentions}) and individual function curves(figure \ref{fig:mep-lineplot}) reveals that the MEP kernel function decays towards zero at a slower rate compared to other kernel functions. Theoretical derivations(see Appendix) have confirmed that the combined $\text{gaussian}+\exp$ kernel function exhibits greater smoothness than either $\text{gaussian}$ or $\exp$ independently. Both experimental evidence and theoretical analyses support the superiority of MEP over current methodologies.
The MKL paper ~\citep{gonen2011multiple} introduces a method for fusing multiple kernel functions. The paper also substantiates the fusion method's effectiveness in harnessing the advantages of individual kernel functions. Utilizing datasets from OpenWebText2, GitHub, and arXiv, our results surpassed those of the ALiBi-type RPE method. Consequently, our approach has been validated as effective.

There are several important limitations of the current method. Given the priority on training speed, a learnable fusion method was not employed. As indicated by the findings in the MKL paper, employing a learnable fusion method could potentially enhance performance further.

Findings from this paper are important. The proposed method employs a straightforward weighted fusion of different kernel functions to enhance positional encoding (PE) effectiveness. Within the realm of parameter learning, our method can seamlessly augment state-of-the-art (SOTA) approaches, thereby improving their performance. By incorporating only a concise snippet of code, current PE performance can be significantly improved.

In conclusion, on multiple datasets, experimental results demonstrate that our method is effective, which uses MKL to fuse multiple kernel functions to generate bias, which is then applied to post-softmax attention scores. Only simple merge kernels, our method outperforms existing approaches in both non-parametric and parametric settings. In practice, for non-parametric approaches, the $\exp$, $\text{gaussian}$, and $\exp(0.5)$ kernel functions can be fused. For parametric approaches, the $\exp+\text{kerple-log}$ kernel function can be fused.

In future work, additional methods such as the T5 model will be integrated to enhance performance for sequence lengths of 512, 1024, and 2048. While prioritizing the speed of training and inference, a learnable approach will be employed for MKL optimization.

\section*{Conclusion}
Current alibi-type relative position coding methods exhibit limitations. With regard to bias generation, these methods employ a singular kernel function, which fails to fully utilize the potential of kernel-based approaches and lacks a mechanism to integrate the benefits of both established and emerging techniques. This paper proposes a novel positional encoding method, called MEP, which utilizes Multi-Kernel Learning (MKL) to improve relative positional encoding in Transformer models. The method effectively fuses multiple kernel functions to generate biases that influence post-softmax attention scores. On datasets from OpenWebText2, GitHub, and arXiv, our MEP results outperformed those of the state-of-the-art (SOTA) method. Despite the promising results, the current method has limitations. Future work should focus on integrating additional methods and employing a learnable MKL optimization approach while considering training and inference speed.

\bibliography{reference}

\section*{APPENDIX}
\label{appendix:exp_gauss_proof}
\paragraph{proof}
To prove that $|k_{MKL}'(x)| < |k_e'(x)|$, we first calculate the derivatives of $k_{MKL}(x)$ and $k_e(x)$.

$k_{MKL}'(x) = 0.5 \cdot k_g'(x) + 0.5 \cdot k_e'(x)$

$k_g'(x) = -\frac{x}{\sigma_1^2} \exp\left(-\frac{x^2}{2\sigma_1^2}\right)$

$k_e'(x) = -\frac{1}{\sigma_2} \exp\left(-\frac{|x|}{\sigma_2}\right) \cdot \text{sign}(x)$

Substituting $k_g'(x)$ and $k_e'(x)$ into $k_{MKL}'(x)$:

$k_{MKL}'(x) = -0.5 \cdot \frac{x}{\sigma_1^2} \exp\left(-\frac{x^2}{2\sigma_1^2}\right) - 0.5 \cdot \frac{1}{\sigma_2} \exp\left(-\frac{|x|}{\sigma_2}\right) \cdot \text{sign}(x)$

Now, we need to prove:

$|k_{MKL}'(x)| < |k_e'(x)|$

$\Leftrightarrow \left|-0.5 \cdot \frac{x}{\sigma_1^2} \exp\left(-\frac{x^2}{2\sigma_1^2}\right) - 0.5 \cdot \frac{1}{\sigma_2} \exp\left(-\frac{|x|}{\sigma_2}\right) \cdot \text{sign}(x)\right| < \left|-\frac{1}{\sigma_2} \exp\left(-\frac{|x|}{\sigma_2}\right) \cdot \text{sign}(x)\right|$

$\Leftrightarrow \left|-0.5 \cdot \frac{x}{\sigma_1^2} \exp\left(-\frac{x^2}{2\sigma_1^2}\right) - 0.5 \cdot \frac{1}{\sigma_2} \exp\left(-\frac{|x|}{\sigma_2}\right) \cdot \text{sign}(x)\right| < \frac{1}{\sigma_2} \exp\left(-\frac{|x|}{\sigma_2}\right)$

Using the triangle inequality:

$\left|-0.5 \cdot \frac{x}{\sigma_1^2} \exp\left(-\frac{x^2}{2\sigma_1^2}\right) - 0.5 \cdot \frac{1}{\sigma_2} \exp\left(-\frac{|x|}{\sigma_2}\right) \cdot \text{sign}(x)\right| \leq 0.5 \cdot \left|\frac{x}{\sigma_1^2} \exp\left(-\frac{x^2}{2\sigma_1^2}\right)\right| + 0.5 \cdot \left|\frac{1}{\sigma_2} \exp\left(-\frac{|x|}{\sigma_2}\right)\right|$

Therefore, if we can prove:

$0.5 \cdot \left|\frac{x}{\sigma_1^2} \exp\left(-\frac{x^2}{2\sigma_1^2}\right)\right| + 0.5 \cdot \left|\frac{1}{\sigma_2} \exp\left(-\frac{|x|}{\sigma_2}\right)\right| < \frac{1}{\sigma_2} \exp\left(-\frac{|x|}{\sigma_2}\right)$

then the original inequality holds.

Simplifying:

$0.5 \cdot \left|\frac{x}{\sigma_1^2} \exp\left(-\frac{x^2}{2\sigma_1^2}\right)\right| < 0.5 \cdot \frac{1}{\sigma_2} \exp\left(-\frac{|x|}{\sigma_2}\right)$

$\Leftrightarrow \left|\frac{x}{\sigma_1^2} \exp\left(-\frac{x^2}{2\sigma_1^2}\right)\right| < \frac{1}{\sigma_2} \exp\left(-\frac{|x|}{\sigma_2}\right)$

$\Leftrightarrow \frac{|x|}{\sigma_1^2} \exp\left(-\frac{x^2}{2\sigma_1^2}\right) < \frac{1}{\sigma_2} \exp\left(-\frac{|x|}{\sigma_2}\right)$

$\Leftrightarrow |x| \cdot \exp\left(-\frac{x^2}{2\sigma_1^2} + \frac{|x|}{\sigma_2}\right) < \frac{\sigma_1^2}{\sigma_2}$

When $|x|$ is sufficiently large, the left side approaches 0, while the right side is a constant. Therefore, there exists an $x_0$ such that for all $|x| > x_0$, the inequality holds.

Thus, we have proved that when $|x|$ is sufficiently large, $|k_{MKL}'(x)| < |k_e'(x)|$ holds.

\paragraph{Model Configurations}
\begin{table}[!ht]
    \centering
    \setlength{\tabcolsep}{3pt}
    \caption{\textbf{162M Model Configurations.}}
    \begin{tabular}{ccccc}
        \hline\hline
         \# Layers & Hidden Size & \# Attention Heads & Train Seq. Len. & \# Trainable Params.\\
         12 & 64 & 12 & 512 & ~162M\\ \hline
         Optimizer & Batch Size & Train Steps & Precision & \# Trainable Params. for RPEs\\
         Adam (lr 6e-4) & 32 & 50,000 & bfloat16 & at most 36\\
         \hline\hline
    \end{tabular}
    \label{tab:model_configs}
\end{table}

\end{document}